\documentclass[runningheads]{llncs}

\usepackage{eccv}

\usepackage{eccvabbrv}

\usepackage{graphicx}
\usepackage{booktabs}
\usepackage{multirow}
\usepackage{multicol}
\usepackage{bbding}

\usepackage[accsupp]{axessibility}  

\usepackage{colortbl}
\usepackage{float}
\usepackage{stfloats}
\usepackage{makecell}

\usepackage[pagebackref,breaklinks,colorlinks]{hyperref}

\usepackage{orcidlink}
\usepackage{wrapfig}

\begin{document}

\title{CAT \includegraphics[width=0.65cm]{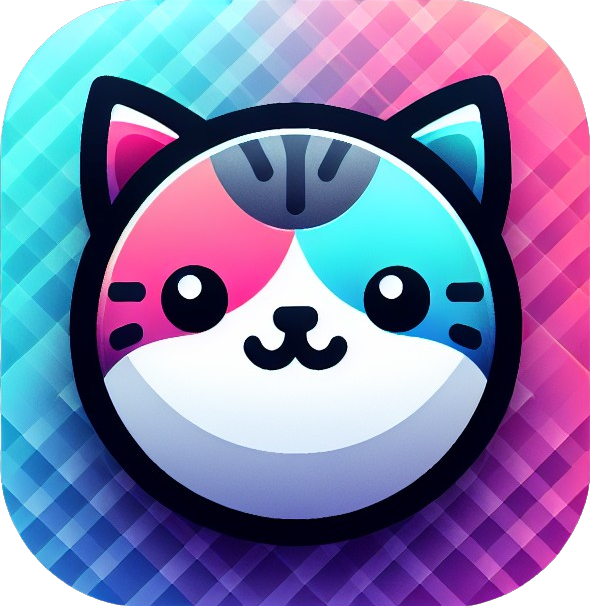}$:$ Enhancing Multimodal Large Language Model to Answer Questions in Dynamic Audio-Visual Scenarios}

\titlerunning{CAT}

\author{
Qilang Ye\inst{1} \and
Zitong Yu\inst{1}\thanks{Corresponding author} \and
Rui Shao\inst{2} \and
Xinyu Xie\inst{1} \and \\
Philip Torr\inst{3} \and
Xiaochun Cao\inst{4}}

\authorrunning{Q.~Ye et al.}

\institute{Great Bay University \and
Harbin Institute of Technology, Shenzhen \and
University of Oxford \and
Shenzhen Campus of Sun Yat-sen University }

\maketitle
\begin{abstract}
  This paper focuses on the challenge of answering questions in scenarios that are composed of rich and complex dynamic audio-visual components. Although existing Multimodal Large Language Models (MLLMs) can respond to audio-visual content, these responses are sometimes ambiguous and fail to describe specific audio-visual events. To overcome this limitation, we introduce the CAT, which enhances MLLM in three ways: 1) besides straightforwardly bridging audio and video, we design a clue aggregator that aggregates question-related clues in dynamic audio-visual scenarios to enrich the detailed knowledge required for large language models. 2) CAT is trained on a mixed multimodal dataset, allowing direct application in audio-visual scenarios. Notably, we collect an audio-visual joint instruction dataset named AVinstruct, to further enhance the capacity of CAT to model cross-semantic correlations. 3) we propose AI-assisted ambiguity-aware direct preference optimization, a strategy specialized in retraining the model to favor the non-ambiguity response and improve the ability to localize specific audio-visual objects. Extensive experimental results demonstrate that CAT outperforms existing methods on multimodal tasks, especially in Audio-Visual Question Answering (AVQA) tasks. The codes and the collected instructions are released at \url{https://github.com/rikeilong/Bay-CAT}.
  
\end{abstract}

\section{Introduction}
\label{sec:intro}
\begin{figure}[t]
  \centering
  \includegraphics[scale=0.3]{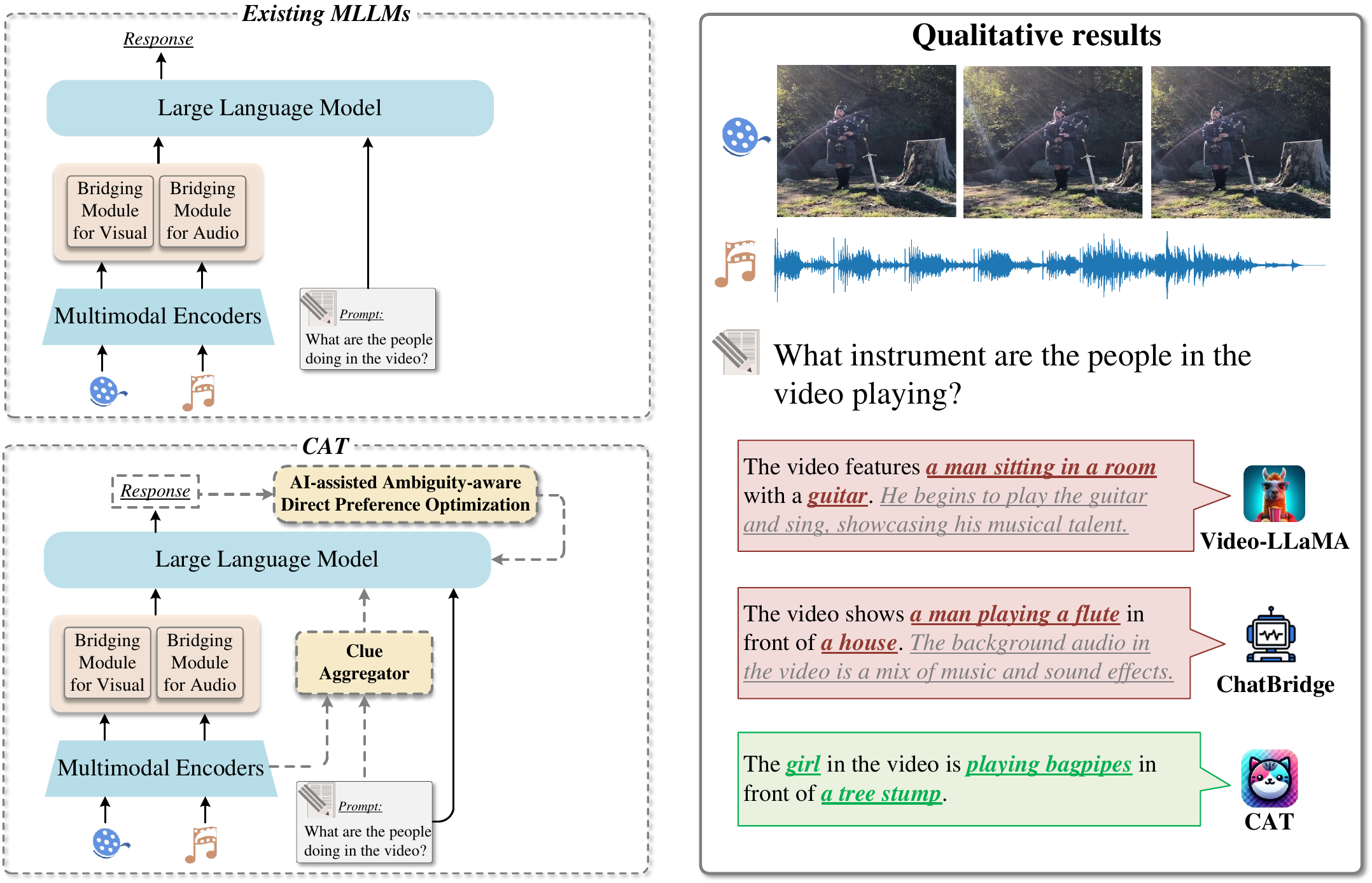}
  \vspace{-0.8em}
  \caption{Comparison between existing MLLMs and CAT \includegraphics[width=0.4cm]{logo3-1.png}. \textcolor[RGB]{146,57,49}{\textbf{Red}} words for incorrect response, \textcolor[RGB]{0,176,80}{\textbf{Green}} words for correct response, and \textcolor{gray}{\textbf{Gray}} words for useless response. \textbf{Left:} Most of the existing MLLMs straightforwardly bridge multimodal to large language models. Instead, CAT builds on this foundation by designing the clue aggregator to learn more detailed knowledge related to the question. Moreover, CAT constrains itself to learn a sharper response through AI-assisted ambiguity-aware direct preference optimization. \textbf{Right:} In comparison with audio-visual-language models Video-LLaMA \cite{videollama} and ChatBridge \cite{chatbridge}, our method accurately recognizes the answers to questions with the most streamlined responses.
  }
   \label{fig1}
   \vspace{-0.8em}
\end{figure}

The real world revolves around sound and visual information, and their combination enhances our ability to perceive the world. Similarly, the development of Multimodal Large Language Models (MLLMs) \cite{videollama,chatbridge,onellm} are closely related to audio and visual. Supervised fine-tuning \cite{sft,rhlfff,VisualInstructionTuning} with specific instruction datasets empowers Large Language Models (LLMs) for multimodal understanding. Despite MLLMs' strong causal reasoning achieving impressive results in common-sense answering \cite{commonqa1,commonqa2}, academic answering \cite{academic}, etc., predicting questions in dynamic audio-visual scenarios remains challenging. This is due to the difficulty of aligning LLMs with cross-domain data during training on large-scale multimodal corpora. It leads to particular ambiguity when describing specific objects in dynamic audio-visual scenarios. These interferences not only cause the model to make random guesses but also seriously affect the inference of detailed answers in Audio-Visual Question Answering (AVQA) \cite{avqa,musicavqa,pianoavqa,avsd} tasks.

A series of bridging modules \cite{vista-llama,chat-univi,llama-vid,monkey,llama-adapterv2} for MLLMs have been designed. The simplest methods \cite{vtimellm,video-llava,videochat,imagebind-llm} use projectors to directly align text with other modalities, but partially limit the ability to capture fine-grained information. In addition, using a cross-attention mechanism to query the audio-visual context \cite{vista-llama,text-conditioned,macaw,instrutblip} to solve multimodal alignment problems is effective but still occurs that a certain visual object or sound cannot be localized in practice.

In this work, we explore possible factors contributing to the above failures: \textbf{1) Audio-visual insufficiently correlates with the question.} As illustrated in the upper left of Fig. \ref{fig1}, most existing MLLMs \cite{videollama,videochat,chatbridge} are mainly designed with multiple branches to handle multiple modalities individually, followed by concatenating the modality embeddings with prompts as inputs to the LLM. This paradigm fails to allow textual information to interact with audio-visual at a low level, resulting in the network's inability to focus on details relevant to the question. \textbf{2) Alignment of multimodal with text is challenging.} The multimodal-text corpus \cite{chatbridge} is difficult to be aligned, making the model generation of responses sometimes ambiguous. This ambiguity generally manifests in ambiguous words of the corresponding audio-visual content, as well as in generating useless text that overly responds. For example, the descriptions illustrated by Video-LLaMA \cite{videollama} and ChatBridge \cite{chatbridge} in Fig. \ref{fig1} are all relatively ambiguous in describing the question-related video and audio content. These descriptions fail to identify the correct answer for ``bagpipes'' when asked for audio-visual reasoning. In addition, when the description is accompanied by many useless and controversy-prone words, it is unfavorable to evaluate the closed-ended AVQA task or open-ended AVQA task.

To overcome the two aforementioned issues, we introduce the CAT, enhancing MLLM in three ways: \textbf{1) Aggregation of question-related key clues.} As illustrated in the lower left of Fig. \ref{fig1}, besides the global visual and audio information, we design a clue aggregator to dynamically capture question-aware visual and audio hidden features to enrich fine-grained clues. The aggregator receives sufficient low-level textual information to interact with audio-visual, it is capable of improving audio-visual grounding. \textbf{2) Mixed audio-visual training strategy.} The training of CAT includes feature alignment using video-text pairs and audio-text pairs, and high-quality instructions to enhance audio-visual awareness. This strategy allows the CAT to be directly involved in real-world scenarios containing both visual and sound. Notably, we collect an audio-visual joint instruction dataset, named AVinstruct, to further empower CAT in AVQA tasks. \textbf{3) Retraining MLLMs to mitigate ambiguity.} The DPO proposed by Rafailov et al. \cite{dpo} has inspired us. Although MLLMs after training are equipped with multimodal understanding, they lack the flexibility to perceive ambiguity. Therefore, we reframe ambiguity elimination as a model preference optimization process and propose an AI-assisted Ambiguity-aware Direct Preference Optimization (ADPO) strategy. Specifically, we refer to ambiguous responses that express the lack of clarity of specific audio-visual objects as negative responses, and then we collect negative responses in the training set and utilize GPT to rewrite them into positive responses. After the multimodal training, we perform ADPO to retrain the model to bias towards the positive response, which is the precise description after the rewrite, and reject the negative response, which is the ambiguous description. Through this learning strategy, CAT can constrain itself to favor non-ambiguity responses. As shown in Fig. \ref{fig1} on the right, CAT correctly responds to the question and excludes all useless information.

Our main contributions are summarized as follows:
\vspace{-0.2em}
\begin{itemize}
    \item We introduce a novel audio-visual-language model, dubbed as CAT, that is capable of learning question-related clues and engaging directly in dynamic audio-visual inference. Notably, we collect AVinstruct, an audio-visual joint instruction dataset to ensure the stability of CAT in AVQA tasks.
    \item As a powerful learning strategy, we propose AI-assisted ambiguity-aware direct preference optimization to overcome the problem that MLLMs tend to ambiguously describe specific audio-visual objects.
    \item We evaluate CAT on a wide range of multimodal tasks. Extensive experiments demonstrate the superiority of CAT (e.g., outperforms the state-of-the-art on a variety of AVQA tasks, and achieves remarkable results in the evaluation of video-based text generation tasks and zero-shot video question-answering tasks).
\end{itemize}

\vspace{-0.8em}
\section{Related Works}
\textbf{Audio-visual question answering (AVQA).} AVQA aims to produce the most accurate linguistic representation of a given video and audio based on the question content, which requires multimodal understanding and reasoning across different semantic levels. The earliest studies \cite{msrvtt-qa,activityqa,vatex} emphasize the understanding of the whole video and return a simple word that is relatively correct to the question. Subsequently, the emergence of dynamic audio-visual datasets (e.g., music scenes \cite{musicavqa,avqa}, and 360-degree panoramas \cite{pianoavqa}) increases the challenge of Question-Answering (QA), which requires the mining of temporal and spatial information in different modalities. For instance, Music-AVQA \cite{musicavqa} requires distinguishing how many instruments are present based on the audio content. AVQA \cite{avqa} requires finding the most plausible of multiple options based on the audio-visual clues from dynamic scenarios.

\vspace{0.4em}
\noindent\textbf{MLLMs for audio-visual question answering.}
Extending LLM to other multimodal tasks is a recently emerging field. Despite the ability of MLLMs to combine information between different modalities, performance on downstream tasks, especially AVQA, remains suboptimal. Many works \cite{instrutblip,chat-univi,vista-llama} emphasize the design of elegant bridging methods to improve the performance of question answering. The simplest bridging modules \cite{videollama,videochatgpt,valley} use one or more linear layers for feature alignment. Although such methods minimize parameter updates, they still have limitations in exploring fine-grained information. Others design more complex bridging networks to query visual information. For example, Lyu et al. \cite{macaw} propose an alignment module for harmonizing different representations before entering the LLM. Ma et al. \cite{vista-llama} propose to keep the distance between all visual tokens and any linguistic tokens consistent within the LLM. However, a large multimodal corpus is difficult to align during training, neither algorithms with low parameter counts \cite{llama-adapterv2} nor complex bridging networks \cite{LION} are prone to the problem of failing to accurately depict specific audio-visual events.

\vspace{0.4em}
\noindent\textbf{Human-preference learning.}
Reinforcement Learning from Human Feedback (RLHF) \cite{rhlfff,rhlf2} is the most classical instance of human preference learning \cite{hpl1,llama2}, it constructs a reward model to optimize the policy model to favor preference responses. Dai et al. \cite{instrutblip} have demonstrated that such preference learning can enhance LLM to generate more accurate information. Recently, Rafailov et al. \cite{dpo} propose a Direct Preference Optimization (DPO) strategy that learns preferences directly by bypassing learning reward models, this simple yet efficient approach inspires us to solve the ambiguity problem.

\section{Our Approach}
\begin{figure}[tb]
  \centering
  \includegraphics[scale=0.268]{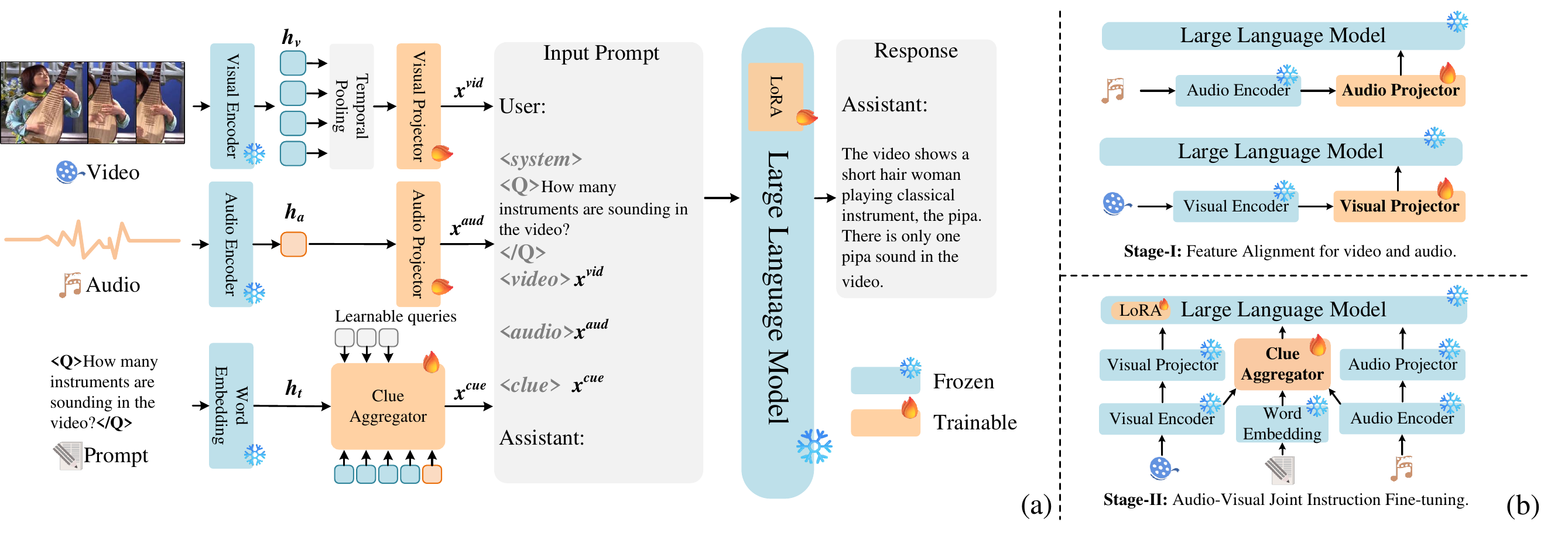}
  \vspace{-1.8em}
    \caption{Illustration of the proposed CAT \includegraphics[width=0.4cm]{logo3-1.png} and its training strategy. (a) Overview of CAT. CAT first extracts overall audio-visual knowledge from video and audio and transforms them into visual tokens $x^{vid}$ and audio tokens $x^{aud}$. We input question tagged with <Q></Q> in the prompt into the clue aggregator, aiming to aggregate question-aware audio-visual hidden features and yield clue tokens $x^{cue}$. Finally, we merge multimodal tokens and language and feed into the frozen large language model with LoRA \cite{lora} to output the response. (b) The training paradigm of CAT involves pre-alignment of the audio-visual projectors and instruction tuning on the entire model.
      }
   \label{mainfig}
\end{figure}

In this section, we first present in detail the proposed CAT (Sec. \ref{sec1} and \ref{sec2}). As shown on the left in Fig. \ref{mainfig}, CAT consists of three branches that draw on visual knowledge, audio knowledge, and question-related clues to feed into the LLM, respectively. Second, we present the multimodel training strategy for CAT as shown on the right in Fig. \ref{mainfig}, where a high-quality audio-visual joint instruction dataset is collected for further fine-tuning (Sec. \ref{sec3}). Lastly, we introduce the AI-assisted Ambiguity-aware Direct Preference Optimization (ADPO) strategy, which reinforces CAT to favor non-ambiguous descriptions (Sec. \ref{sec4}).

\vspace{-0.8em}
\subsection{Multimodal Inputs} \label{sec1}
ImageBind \cite{imagebind} with a single joint embedding space demonstrates strong modality learning ability. Therefore, we leverage the frozen ImageBind as a universal encoder for all modalities. Benefiting from the robust pre-alignment of both visual and audio to text in the encoder, CAT can easily transfer cross-domain knowledge. Given a video $\textbf{V}$ and an audio $\textbf{A}$, the encoded multimodal hidden features can be obtained by:
\begin{equation}\label{eq:softmask}
h_{v}=\text{ImageBind}(\textbf{V}), \quad h_{a}=\text{ImageBind}(\textbf{A}),
\end{equation}
where $h_{v} \in {\mathbb{R}^{T \times d_{h}}}$, $h_{a} \in {\mathbb{R}^{1 \times d_{h}}}$ are video and audio features, respectively. $T$ denotes the length of the given video and $d_{h}$ is the specific dimension. Notably, we temporally compress the frame-level features $h_{v}$ to address the computational burden. Further, we employ two linear projection layers to align the hidden dimensions of the inputs $h_{v}$, $h_{a}$ to obtain visual tokens $x^{vid}$ and audio tokens $x^{aud}$, respectively.

\vspace{-1em}
\subsection{Aggregating Key Clues}\label{sec2}
\begin{wrapfigure}[15]{l}{6cm} 
\includegraphics[scale=0.27]{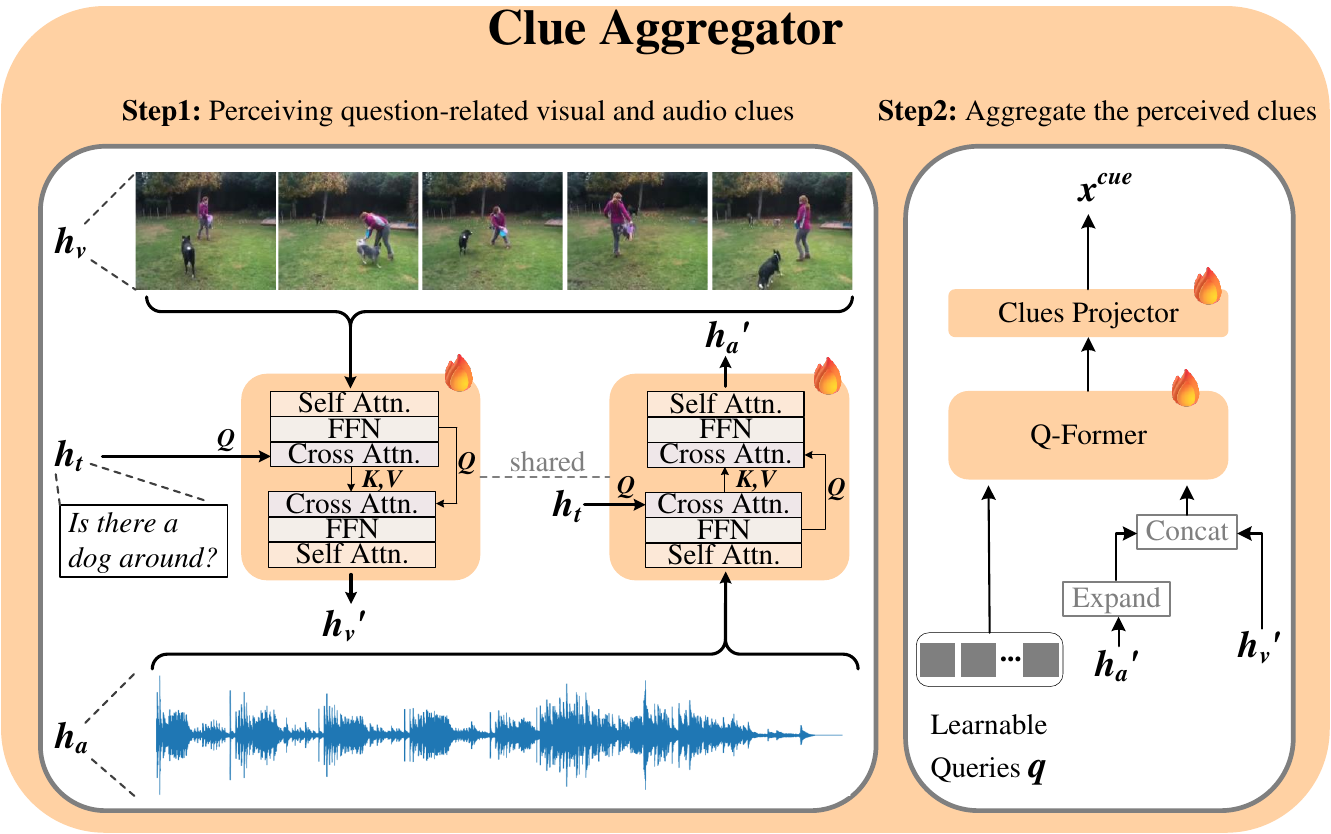}\
\caption{Illustration of clue aggregator in a simple example.}\label{fig:CA}
\end{wrapfigure}

An important component of MLLMs is the design of efficient bridging modules. The simple approach is to apply linear layers to incorporate the full visual and audio. The advantage is that it does not require any reduction of spatio-temporal information. However, such bridge approaches sometimes make LLM fail to reflect certain content and consistently generate a chunk of description about the video \cite{videollama,videochat}. Therefore, we devise a Clue Aggregator (CA) that enriches LLM-acquired knowledge by mining multimodal clues related to the question. CA is divided into two steps, as illustrated in Fig. \ref{fig:CA}.  

\vspace{0.2em}
\noindent\textbf{Step1: perceiving question-related visual and audio clues.} To extract more sufficient details from the input video and audio, we devise a perceiver, which consists of multiple different transformer layers. The perceiver can be viewed as a tiny transformer network that arranges the self-attention (SA), cross-attention (XA), and feed-forward network (FFN) in forward and reverse order. The forward order block $\mathcal{B}_1$ is to perform attention-based question-aware localization. The reverse order block $\mathcal{B}_2$ is to consolidate the original audio-visual representation at the attentional level. Specifically, given tokens representing ``\textit{Is there a dog around?}'' and the corresponding text embedding $h_{t}$:
\begin{equation}\label{eq3}
\mathcal{B}_1(h_t;X) = \text{XA}\left(h_t, \text{FFN}(\text{SA}(X)) \right),
\end{equation}
\vspace{-0.8em}
\begin{equation}\label{eq4}
\mathcal{B}_2(X;\mathcal{B}_1(h_t;X)) = \text{SA}(\text{FFN}(\text{XA}(\text{FFN}(\text{SA}(X)), \mathcal{B}_1(h_t,X)))).
\end{equation}
In XA($\cdot$,$\cdot$), the former represents the query and the latter represents the key and value. Based on Eqs. \ref{eq3} and \ref{eq4}, we use $h_{v}'=\mathcal{B}_2(h_{v};\mathcal{B}_1(h_t;h_{v}))$ and $h_{a}'=\mathcal{B}_2(h_{a};\mathcal{B}_1(h_t; h_{a}))$ to obtain $h_{v}'$, $h_{a}'$, the question-aware visual and audio features, respectively. Notably, the perceiver for visual and the perceiver for audio implement shared parameters to learn potential associations.

\vspace{0.2em}
\noindent\textbf{Step2: aggregating the perceived clues.} A three-minute video undergoes a frozen encoder to get frame-level features of about 400 to 500 in length, leading to the fact that general machines cannot bear the burden of feeding all the frame-level features into the LLM. Thanks to the Q-former architecture proposed by Li et al. \cite{blip2}, which reduces the computational cost of end-to-end training of multimodal-language models. Specifically, we customize a learnable query vector $q$ in length $K$ to further extract useful information from the input question-aware features. Notably, we expand the $h_{a}'$ in the time dimension to be consistent with $h_{v}'$ and concatenate them. In practice, we set $K=48$ to aggregate all question-aware features and obtain the clue tokens $x^{cue}$ via a clues projector to align the dimension with the LLM.

\vspace{-0.8em}
\subsection{Multimodal Training Strategy} \label{sec3}
\vspace{-0.1em}
We follow the previous works \cite{VisualInstructionTuning} to promote comprehension over video and audio. As shown in Fig. \ref{mainfig} on the right, we pre-training CAT on the video-level and audio-level tasks in stage-\uppercase\expandafter{\romannumeral1}. In stage-\uppercase\expandafter{\romannumeral2}, we fine-tune CAT on high-quality audio-visual-level instructions.

\vspace{0.2em}
\noindent\textbf{Stage-\uppercase\expandafter{\romannumeral1}: feature alignment for video and audio projectors.} First, we employ the video-text Webvid 2.5M dataset \cite{webvid} to train the visual projector (audio projector not involved in training). Second, we employ the audio-text WavCap dataset \cite{wavcap} to train the audio projector (visual projector not involved in training). During the above two training periods, we freeze the LLM and the encoder to align the semantic information for vision and audio, respectively.

\vspace{0.2em}
\noindent\textbf{Stage-\uppercase\expandafter{\romannumeral2}: audio-visual joint instruction tuning.} To equip CAT with the ability to reason jointly based on visual and auditory, we collect an audio-visual joint instruction dataset, named AVinstruct, which emphasizes co-learning dynamic audio-visual pairs to solve diverse AVQA tasks. Specifically, we collect a large number of raw videos containing audio information from YouTube and VGGSound \cite{vggsound} as well as QA pairs from the training set in closed-ended AVQA tasks (i.e. music scene \cite{musicavqa} and real-world scene \cite{avqa}). Then different subtitle generators (i.e. BLIP2 \cite{blip2} and Whisper \cite{whisper}) are utilized to obtain video descriptions. Finally, we use GPT to generalize from human-written examples to synthesize question-guided audio-visual descriptions. We integrate all compositions after stage-\uppercase\expandafter{\romannumeral1} training and freeze the visual projector and audio projector, only fine-tuning clue aggregator and combined LoRA parameters on 100k video instruction \cite{videochatgpt} and AVinstruct. Notably, to highlight the question for applying in the clue aggregator, we reconstruct the form of the input prompt by adding two simple tokens $<Q>$, $</Q>$, where $<Q>$ denotes the start position of the question and $</Q>$ denotes the end position of the question. During the instruction fine-tuning phase, we use predefined prompts based on the following template:
\begin{equation*}
\text{USER}:<system><Q></Q><video><audio><clues> \quad \text{Assistant}:
\end{equation*}
In this prompt, $<system>$ denotes the official guidance message. $<video>$, $<audio>$, and $<clues>$ are associated with visual tokens, audio tokens, and question-related clue tokens, respectively.

\vspace{-1em}
\subsection{AI-assisted Ambiguity-aware Direct Preference Optimization}\label{sec4}
At first, we attempt to design various prompts to allow MLLMs to generate the most concise descriptions. However, such a no-learning approach is challenging to achieve the goal of accurate responses in a variety of dynamic audio-visual scenarios. We find that MLLMs are prone to ambiguity when describing specific audio-visual objects. For example, an ambiguous response is illustrated by trained-CAT in Fig. \ref{fig4}, where trained-CAT denotes CAT after feature alignment and instruction tuning. ADPO is designed to allow MLLMs to review their mistakes and relearn with a new objective to better expression. Specifically, ADPO is divided into two steps to optimize CAT:

\begin{figure}[tb]
  \centering
  \includegraphics[scale=0.2]{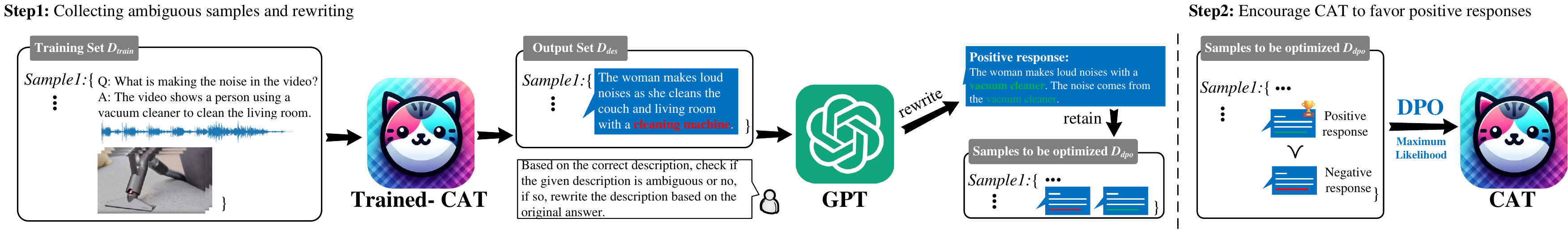}

  \caption{Trained-CAT denotes CAT after feature alignment and instruction tuning. Our proposed ADPO strategy involves two steps. First, we collect the negative response generated by trained-CAT and correct it by GPT to obtain a positive response based on the original answer. Second, we perform ADPO training to skew CAT toward positive responses and reject negative responses.
  }
   \label{fig4}
   \vspace{-0.8em}
\end{figure}

\noindent\textbf{Collecting ambiguous samples and rewriting.} As illustrated on the left in Fig. \ref{fig4}, assume that a given training set $D_{train}$ and the corresponding trained-CAT output set $D_{des}$. We first provide GPT with a detailed prompt template to review each output in $D_{des}$ for ambiguity. The rule of the review is to identify output ambiguities, i.e., responses that differ significantly from the original answer, which is referred to as a negative response. We let GPT correct this negative response, with the principle of generating a positive response that clearly describes the visual objects or sound without significant modification. Repeat the above steps to get the sample set $D_{dpo}$ to be optimized, where $D_{dpo}=\{x^{vid},x^{aud},x^{cue},x^{txt},y_{pos},y_{neg}\}$. $x$ denotes the input modal token and $x^{txt}$ denotes the text tokens,  $y_{pos}$, $y_{neg}$ denote the positive and negative responses, respectively.

\vspace{0.2em}
\noindent\textbf{Encouraging CAT to favor positive responses.} As illustrated on the right in Fig. \ref{fig4}, ADPO directly optimizes the low-rank adaptation matrix parameters \cite{lora} in the policy model. We assume CAT as a policy model $f^{pol}$, given a reference model $f^{ref}$, a positive response $y_{pos}$, and a negative response $y_{neg}$. The reference model $f^{ref}$ is a deep copy of $f^{pol}$ and during training we encourage $f^{pol}$ to favor $y_{pos}$, while $f^{ref}$ does not update the weights. Specifically, we define DPO loss $\mathcal{L}_{DPO}$ as:
\begin{equation}\label{dpo}
\begin{split}
\mathcal{L}_{DPO}\left(f^{pol};f^{ref}\right) = & -E_{\left(x^{vid},x^{aud},x^{cue},x^{txt},y_{pos},y_{neg}\right)\sim D_{dpo}} \\
&\bigg[ log \sigma \bigg( \beta log\frac{f^{pol}(y_{pos}|\left[x^{vid}:x^{aud}:x^{cue}:x^{txt}\right])}{f^{ref}(y_{pos}|\left[x^{vid}:x^{aud}:x^{cue}:x^{txt}\right])} \\
&- \beta log\frac{f^{pol}(y_{neg}|\left[x^{vid}:x^{aud}:x^{cue}:x^{txt}\right])}{f^{ref}(y_{neg}|\left[x^{vid}:x^{aud}:x^{cue}:x^{txt}\right])}\bigg) \bigg],
\end{split}
\end{equation}
$[· \quad : \quad ·]$ denotes the concatenation. $\sigma$ is the non-linear function, Sigmoid, and $\beta$ is a hyperparameter. This objective function aims to directly optimize the model that favors the positive response $y_{pos}$ and rejects the negative response $y_{neg}$. Assuming that $\beta=1$, we simplify the output $\hat{r}$ of the model based on Eq. \ref{dpo} as:
\begin{equation}
\hat{r}(x^{vid},x^{aud},x^{cue},x^{txt},y) = log\frac{f^{pol}(y|\left[x^{vid}:x^{aud}:x^{cue}:x^{txt}\right])}{f^{ref}(y|\left[x^{vid}:x^{aud}:x^{cue}:x^{txt}\right])}.
\end{equation}
The optimal goal is to make the variance of $\hat{r}(x^{vid},x^{aud},x^{cue},x^{txt},y_{pos})-\hat{r}(x^{vid},$ $x^{aud},x^{cue},x^{txt},y_{neg})$ larger and larger as a way to steer the model to generate descriptions devoid of ambiguous. However, when the difference between positive and negative responses is small, DPO loss supervision alone is not an effective facilitator. Therefore, we introduce an additional objective $\mathcal{L}_{SFT}$, which is similar to the process of supervised fine-tuning, to supervise the model to stabilize the bias towards the positive response. Specifically, we define $\mathcal{L}_{SFT}$ as:
\begin{equation}
\begin{split}
\mathcal{L}_{SFT} = - \sum \text{log}P(y_{pos}|\left[x^{video}:x^{audio}:x^{clues}:x^{txt}\right];f^{pol}), \\
\{x^{vid},x^{aud},x^{cue},x^{txt},y_{pos}\}\sim D_{dpo},
\end{split}
\end{equation}
this loss function can be interpreted as using the positive output probability in the policy model to aid in optimization. During ADPO training, we sum up $\mathcal{L}_{DPO}$ and $\mathcal{L}_{SFT}$ with a bias $\lambda$ to achieve direct preference optimization:
\begin{equation}
\mathcal{L} = \mathcal{L}_{DPO} + \lambda \mathcal{L}_{SFT},
\end{equation}
where we set $\lambda=0.1$ by default.

\section{Experiments}
\subsection{Datasets}
\noindent\textbf{Video-based text generation tasks.} We evaluate the proposed CAT on the video understanding benchmarks proposed by Video-ChatGPT \cite{videochatgpt}. Specifically, Video-ChatGPT proposes five metrics: \textit{Correctness of Information}, \textit{Consistency}, \textit{Detail Orientation}, \textit{Contextual Understanding}, and \textit{Temporal Understanding}, which test the ability of MLLMs to describe videos.

\vspace{0.2em}
\noindent\textbf{Zero-shot on video question answering tasks.} To evaluate whether CAT has the basic ability to communicate regarding video, we conduct zero-shot tests on MSRVTT-QA \cite{msrvtt-qa} and ActiviytNet-QA \cite{activityqa}. MSRVTT-QA and ActiviytNet-QA consist of 10k and 5.8k videos containing audio information, respectively, where the QA pairs are mostly questions about daily life.

\vspace{0.2em}
\noindent\textbf{Closed-ended AVQA tasks.} We categorize the Music-AVQA \cite{musicavqa} and AVQA \cite{avqa} datasets as closed-ended AVQA tasks. These datasets consist of up to 42 candidate answers that require the selection of the most appropriate one based on visual and auditory content.

\vspace{0.2em}
\noindent\textbf{Open-ended AVQA tasks.} We select audio-visual dialogue (AVSD \cite{avsd}), and audio-visual captioning (VALOR \cite{valor}) tasks for the evaluation of open-ended AVQA. These tasks require precise language to interpret, correlate, and reason about cross-modal information. We evaluate the zero-shot ability of CAT on these datasets.
\vspace{-1.2em}
\subsection{Experimental Setup}
\noindent\textbf{Evaluation metrics.} For video-based text generation and zero-shot on video question-answering tasks, we follow the evaluation pipeline proposed by Video-ChatGPT \cite{videochatgpt}, which is based on GPT-3.5 to evaluate predictive descriptions against correct descriptions, with a score from 0 to 5 indicating accuracy. In this paper, we standardize 0 to 5 as 0 to 100 to align common accuracy rubrics. For closed-ended AVQA tasks, we report the accuracy of the correct sample. For open-ended AVQA tasks, we report the CIDEr \cite{cider} that specializes in evaluating visually descriptive tasks.

\vspace{0.2em}
\noindent\textbf{Architecture.} We use frozen ImageBind \cite{imagebind} and LLaMA2-7B \cite{llama2} as audio-visual encoders and LLM, respectively. The size of modality embeddings for each modality are $\mathbb{R}^{T \times 1024}$. The outputs $x^{vid}$, $x^{aud}$, and $x^{cue}$ are $\mathbb{R}^{1 \times 4096}$, $\mathbb{R}^{1 \times 4096}$, and $\mathbb{R}^{48 \times 4096}$, respectively.

\vspace{0.2em}
\noindent\textbf{Training details.} We complete feature alignment training, instruction tuning, and ADPO training with 1 NVIDIA A100 GPU. In detail, For the feature alignment training and instruction tuning, we use the AdamW optimizer with a cosine learning rate decay and a warm-up period. When LoRA is added, we set r = 64 and alpha = 128 for the LoRA parameters, and the total batch size is set to 128 for training 1 epoch with a learning rate of $2e^{-5}$. For ADPO training, we only select the training set in the instruction tuning phase for optimization. We set r = 64 and alpha = 16 for the LoRA parameters, and the total batch size is set to 1 for training 1 epoch with a learning rate of $4e^{-6}$. The hyperparameter $\beta$ is set to 0.1.

\begin{table}[tb]
  \caption{GPT-based evaluation \cite{videochatgpt} for video-based text generation and zero-shot video question answering. For clarity, five scores are reported ("Cr.": Correctness of Information, "Cs.": Consistency, "De.": Detail Orientation, "Ct": Contextual Understanding, "Te.": Temporal Understanding). Score indicates the confidence.
  }
  \label{tab:headings}
  \centering
  \setlength{\tabcolsep}{1mm}{\scalebox{0.88}{
  \begin{tabular}{@{}l|c|ccccc|cc|cc@{}}
    \hline
    \multirow{2}{*}{Method} & \multirow{2}{*}{LLM Size}  & \multicolumn{5}{c|}{Video-based text generation} & \multicolumn{2}{c|}{\makecell[c]{Zero-shot on \\ MSRVTT-QA}}& \multicolumn{2}{c}{\makecell[c]{Zero-shot on \\ ActivityNet-QA}}\\
     &   & Cr.  & De. & Ct. & Te. & Cs. & Acc. & Score & Acc. & Score \\
     \hline
     LLaMA-VID \cite{llama-vid} & 13B & 61.4 & 61.0 & 72.0  & 51.6 & 52.6 & 58.9 & 3.3 & 47.5 & 3.3\\
    \hline
    Video-LLaMA \cite{videollama} & 7B & 39.2 & 43.6 & 43.2 & 36.4 & 35.8  & 29.6 & 1.8 & 12.4 & 1.1\\
    LLaMA-Adapter \cite{LLaMA-Adapter} & 7B & 40.6 & 46.4 & 46.0 & 39.6 & 43.0 & 43.8 & 2.7 & 34.2 & 2.7\\
    VideoChat \cite{videochat} & 7B & 44.6 & 50.0 & 50.6 & 38.8 & 44.8 & 45.0 & 2.5 & 26.5 & 2.2\\
    Video-ChatGPT \cite{videochatgpt} & 7B & 48.0 & 50.4 & 52.4 & 39.6 & 47.4 & 49.3 & 2.8 & 35.2 & 2.7\\
    VISTA-LLaMA \cite{vista-llama} & 7B & 48.8 & 52.8 & 63.6 & 45.2 & 46.2 & 60.5 & 3.3 & 48.3 & 3.3 \\
    VideoChat2 \cite{videochat2} & 7B & 60.4 & 57.8 & 70.2 & 53.2 & 56.2 & 54.1 & 3.3 & 49.1 & 3.3\\
    Chat-UniVi \cite{chat-univi} & 7B & 57.8 & 58.2 & 69.2 & \textbf{57.8} & 56.2 & 54.6 & 3.1 & 45.8 & 3.2 \\
    LLaMA-VID \cite{llama-vid} & 7B & 59.2 & 60.0 & \textbf{70.6}  & 49.2 & 50.2 & 57.7 & 3.2 & 47.1 & 3.3\\
   \rowcolor{lightgray!30}\textbf{CAT (Ours)} & 7B & \textbf{61.6} & \textbf{62.0} & \underline{69.8} & \underline{56.2} & \textbf{57.8} & \textbf{62.1} & \textbf{3.5} & \textbf{50.2} & \textbf{3.5} \\
  \hline
  \end{tabular} \label{tab1}  }}
\end{table}

\begin{table}[]\centering
\caption{Comparison with non-LLMs-based methods on fine-tuned Music-AVQA dataset.}\label{MUSIC-AVQA}
\vspace{-0.8em}
\setlength{\tabcolsep}{1mm}{\scalebox{0.8}{
\begin{tabular}{l|c|c|c|c|c|c}
\hline
Method & \makecell[c]{Language \\ Model} & \makecell[c]{Trainable \\ Params (M)} & Audio avg. & Visual avg. & \makecell[c]{Audio-Visual \\ avg.}& Overall avg. \\
\hline
FCNLSTM \cite{FCNLSTM} & CLIP \cite{clip} & 9.79 & 68.9& 56.2& 60.4& 60.8 \\
GRU \cite{GRU} & CLIP  & - & 68.3& 67.0& 63.0& 65.0\\
HCAttn \cite{HCAttn} & CLIP  & - & 64.9& 65.3& 60.3& 62.5 \\
MCAN \cite{MCAN} & CLIP  & 56.0 & 70.6& 71.8& 61.5& 65.8 \\
PSAC \cite{PSAC} & CLIP  & - & 72.0& 69.4& 63.6& 66.6  \\
HME \cite{HME}  & CLIP  & - & 69.9& 68.8& 64.8& 66.7  \\
HCRN \cite{HCRN}  & CLIP  & - & 63.7& 65.2& 49.8& 56.3 \\
AVSD \cite{AVSD2} & CLIP  & 8.35 & 68.8& 70.3& 65.4& 67.3  \\
Panp-AVQA \cite{pianoavqa} & CLIP  & - & 72.1& 73.2& 67.0& 69.5  \\
AVST \cite{musicavqa} & CLIP  & 18.48 & 73.9& 74.4& 69.5& 71.6  \\
PSTP-Net \cite{pstp-net} & BERT \cite{bert} & 4.297 & 70.9& 77.3& 72.6& 73.5 \\
LAVISH \cite{LAVISH} & CLIP & 21.09 &$77.1$&$77.3$&$77.0$&- \\
CAD \cite{CAD} & GLoVE \cite{GLoVE} & - &$78.1$&$79.7$&$76.9$&$78.2$ \\
VALOR \cite{valor} & BERT  & - & - & - &-&$78.9$\\
VAST \cite{vast} & BERT  & - & - & - &-&$80.7$\\
\hline
    \rowcolor{lightgray!30}\textbf{CAT-7B (Ours)} & LLaMA2 & 5.813 & \textbf{84.9}&\textbf{86.1}&\textbf{83.2}&\textbf{84.3}  \\
\hline
\end{tabular} \label{tab2}}}
\vspace{-0.8em}
\end{table}

\vspace{-1.2em}
\subsection{Comparison to State-of-the-Art}
 \vspace{-0.2em}
\noindent\textbf{Comparison on video-based text generation tasks.} We follow the benchmark proposed by Video-ChatGPT \cite{videochatgpt} to evaluate CAT. On the left of Table \ref{tab1}, we show that CAT achieves the state-of-the-art in terms of correctness of description information (Cr.), detailed description of the problem (De.), and coherence of the description (Cs.). In addition, CAT does not parse longer visual tokens, it still achieves competitive results when comparing time-related descriptions (e.g., contextual and temporal understanding).

\vspace{0.2em}
\noindent\textbf{Comparison on zero-shot video question answering tasks.} On the right of Table \ref{tab1}, we show the zero-shot video question answering performance of CAT on several open-ended datasets. While recent MLLMs designed with bridging modules have produced substantial results, CAT is way ahead of them in recognizing accurate answers. We consistently outperform the state-of-the-art on the MSRVTT-QA \cite{msrvtt-qa} and ActivityNet-QA \cite{activityqa} benchmarks.

\vspace{0.2em}
\noindent\textbf{Comparison on closed-ended AVQA tasks.} We choose to evaluate Music-AVQA \cite{musicavqa} to demonstrate that CAT is capable of perceiving specific audio-visual objects to answer questions. In Table \ref{tab2}, under full supervision of the training set, CAT accurately retrieves specific objects in dynamic audio-visual scenarios and comprehensively outperforms all non-LLMs-based models. Thanks to LoRA \cite{lora}, we can improve the evaluation quality with the help of the world knowledge inside LLaMA2 \cite{llama2} and only 5.8M parameters are trainable. Also, we examine the ability of CAT for zero-shot on Music-AVQA \cite{musicavqa}. In Table \ref{tab3}, we show the results comparing LLMs-based models. Notably, for a fair comparison, we remove the LoRA parameters of CAT that are fine-tuned on AVinstruct, which is derived from the Music-AVQA \cite{musicavqa} and AVQA \cite{avqa} training sets. Even though our model does not draw on the larger-scale LLM, it still achieves a small advantage over ChatBridge \cite{chatbridge} with a 13B LLM size. Furthermore, we show the performance of CAT in multiple-choice scenarios \cite{avqa} in Table \ref{tab4}, CAT continues outstanding.

\begin{table}[t]
\begin{minipage}{0.4\linewidth}
  \centering
\caption{Comparison with LLMs-based methods on zero-shot Music-AVQA dataset.}
 \scalebox{0.8}{
  \begin{tabular}{@{}lcc@{}}
    \hline
    Method & zero-shot  & Acc.\\
    \hline
    OneLLM-7B \cite{onellm} & \checkmark & 43.0\\
    ChatBridge-13B \cite{chatbridge} & \checkmark & 47.6\\
    \rowcolor{lightgray!30}\textbf{CAT-7B (Ours)} & \checkmark & \textbf{48.6} \\
  \hline
  \end{tabular}\label{tab3}}
  \vspace{+1.5em}
\caption{Evaluation on fine-tuned AVQA dataset.}
 \setlength{\tabcolsep}{1.5mm}{\scalebox{0.8}{
   \begin{tabular}{@{}lc@{}}
   \hline
    Method  & Acc.\\
    \hline
    HGA + HAVF \cite{HGA}  & 87.7\\
    HCRN + HAVF \cite{avqa}  & 89.0\\
    PSTP-Net \cite{pstp-net} & 90.2 \\
    \rowcolor{lightgray!30}\textbf{CAT-7B (Ours)} & \textbf{92.0} \\
  \hline
  \end{tabular}\label{tab4}  }}
\end{minipage}
\begin{minipage}{0.6\linewidth}
\centering
\centering
\caption{Evaluation on zero-shot open-ended AVQA datasets.}
  \setlength{\tabcolsep}{1.5mm}{\scalebox{1}{
  \begin{tabular}{@{}lccc@{}}
    \hline
    Method & zero-shot & \makecell[c]{AVSD \\ CIDEr} & \makecell[c]{VALOR \\ CIDEr}\\
    \hline
    VALOR \cite{valor} & $\times$  & - & 61.5 \\
    VAST \cite{vast} & $\times$ & - & 62.2 \\
    FA+HRED \cite{FA+HRED} & $\times$ & 84.3 & -\\
    MTN \cite{MTN}& $\times$ & 98.5 & -\\
    COST \cite{COST} & $\times$ & 108.5 & -\\
    \hline
    OneLLM-7B \cite{onellm}& \checkmark  & $74.5$ & 29.2 \\
    ChatBridge-13B \cite{chatbridge}& \checkmark & $75.4$  & 24.7 \\
    \rowcolor{lightgray!30}\textbf{CAT-7B (Ours)} &\checkmark & \textbf{79.0}  & \textbf{32.4} \\
\hline
  \end{tabular} \label{tab5}  }}
\end{minipage}
\end{table}

\noindent\textbf{Comparison on open-ended AVQA tasks.} Open-ended AVQA tasks require responses to daily events based on audio-visual content. We conduct a comparative analysis of our model with multimodal-based LLMs: OneLLM \cite{onellm}, and ChatBridge \cite{chatbridge}. The tests are divided into zero-shot and full supervision, and we only choose to evaluate the zero-shot to demonstrate the level of practical application of the CAT. In Table \ref{tab5}, CAT surpasses OneLLM and ChatBridge on zero-shot complex text reasoning tasks AVSD \cite{avsd} and VALOR \cite{valor}. Moreover, our approach is close to the results of FA \cite{FA+HRED} using fully supervised training, further demonstrating the strong multimodal understanding of CAT.

\vspace{-1.2em}
\subsection{Ablations and Analyses}
In this subsection, we explore the effects of each component of CAT. We test the effect of input modal tokens on video-based text generation \cite{videochatgpt}. In addition, to avoid the huge expense associated with the GPT3.5 evaluation, we test each component of the clue aggregator on the test set of AVinstruct, as well as the generalizability of ADPO on AVSD \cite{avsd}.

\vspace{0.2em}
\noindent\textbf{Does question-related clues matter?} Visual tokens and audio tokens already have the ability to distinguish video content. To explore whether the added question-related clues are meaningful, we conduct ablation experiments on the video-based text generation task in Table \ref{tab6}. We first test visual knowledge separately from question-related clues knowledge on video comprehension. Observations reveal that the results achieved when inputting question-related clues alone have been able to outperform the joint input of audio-visual knowledge. Further, we add clues knowledge to the original audio-visual knowledge, and the results demonstrate that the fusion of the three can fully satisfy the information required for LLM reasoning. 

\vspace{0.2em}
\noindent\textbf{Ablation on clue aggregator.} We categorize a portion of the collected audio-visual joint instruction data into a test set and present the ablation results in Table \ref{tab7}. The evaluation indicators B, M, and C denote BLEU-4 \cite{bleu}, METEOR \cite{METEOR}, and ROUGE-L, respectively. We split $\mathcal{B}_1$ and $\mathcal{B}_2$ and query subjects $h_v$ and $h_a$ to explore the impact of CA internals on reasoning about dynamic audio-visual pairs. We find that visual attention influences context much more than auditory. This is because vision has more information that can be mined and is more relevant to the question. Next, we reduce parameters to investigate whether the same performance can be achieved with a single block. However, since the question acts as a query, without $\mathcal{B}_2$ to recover the original modal length does not provide an advantage over two blocks.

\begin{table} [t]
\begin{minipage}{0.35\linewidth}
\centering
\caption{Ablation experiment about input modals on video-based text generation.}
 \scalebox{0.7}{
  \setlength{\tabcolsep}{1.5mm}{
  \begin{tabular}{@{}cccccc@{}}
    \toprule
     \multicolumn{3}{c}{Input-modal} & \multirow{2}{*}{Cr.}  & \multirow{2}{*}{De.} & \multirow{2}{*}{Te.}\\
     $x^{vid}$ &$x^{aud}$ &$x^{cue}$ & & &  \\
    \midrule
   \checkmark & & & 38.6 & 40.4 & 34.0   \\
  \checkmark & \checkmark & & 40.6 & 44.8 & 35.2   \\
   & & \checkmark  & 58.6 & 57.4 & 55.8 \\
   \checkmark & \checkmark & \checkmark & \textbf{61.6} & \textbf{59.0} & \textbf{56.2}  \\
  \bottomrule
  \end{tabular} \label{tab6} }  }
\end{minipage}
\begin{minipage}{0.3\linewidth}
\centering
\caption{Ablation experiments about clue aggregator on audio-visual joint instruction dataset.}
 \scalebox{0.7}{ \setlength{\tabcolsep}{1mm}{
\begin{tabular}{@{}lccc@{}}
    \toprule
    Method & B@4 & M  &  R\\
    \midrule
    $\mathcal{B}_1$ + $\mathcal{B}_2$  & \textbf{30.8} & \textbf{58.7}  & \textbf{68.5} \\
    $\mathcal{B}_1$ + $\mathcal{B}_2$ w/o $h_{v}$  & $10.0$ & $37.1$  & $49.9$ \\
    $\mathcal{B}_1$ + $\mathcal{B}_2$ w/o $h_{a}$ & $18.8$ & $49.3$  & $61.8$ \\
    only $\mathcal{B}_1$ & $25.9$ & $54.7$  & $64.2$ \\
  \bottomrule
  \end{tabular} \label{tab7} }}
\end{minipage}
\begin{minipage}{0.3\linewidth}
\centering
\caption{Ablation experiment about ADPO on AVSD \cite{avsd}.}
 \scalebox{0.7}{
  \setlength{\tabcolsep}{1mm}{
  \begin{tabular}{@{}lcccc@{}}
    \toprule
    \multirow{2}{*}{Model} & \multirow{2}{*}{ADPO}  & \multicolumn{3}{c}{AVSD}\\
    &  & B@4 & M  & C \\
    \midrule
    Video-LLaMA & $\times$ & 18.4 & 40.1 & 63.7 \\
    Video-LLaMA & \checkmark & 22.2 & 48.2 & 69.2 \\
    \textbf{CAT (Ours)} & $\times$ & 28.9 & 56.2 & 74.8 \\
   \textbf{CAT (Ours)} & \checkmark & \textbf{34.2} & \textbf{59.8} & \textbf{79.0} \\
  \bottomrule
  \end{tabular} \label{tab8}  }}
\end{minipage}
 \vspace{-0.8em}
\end{table}

\begin{figure}[tb]
  \centering
  \begin{subfigure}{0.5\linewidth}
    \includegraphics[scale=0.26]{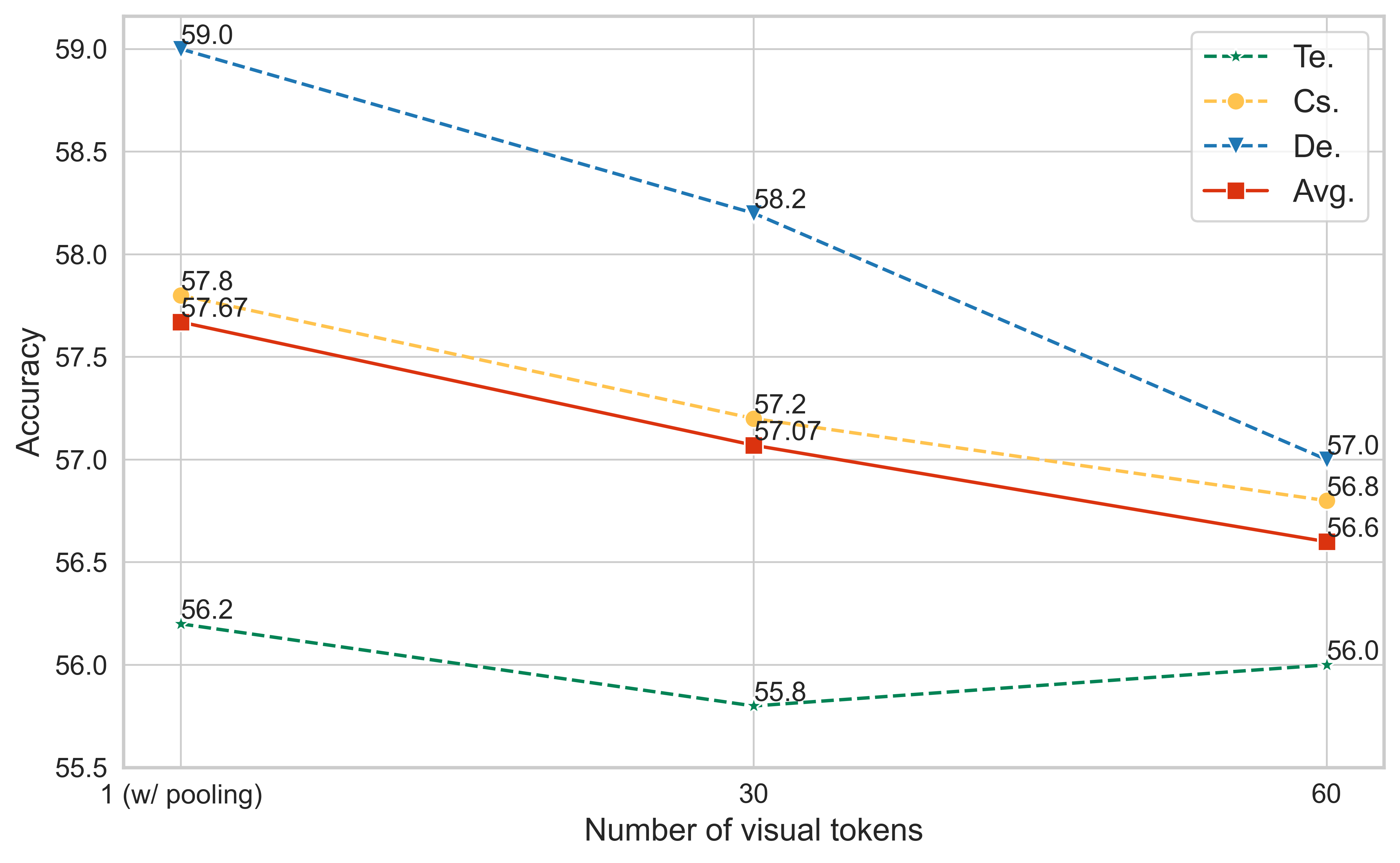}
    \caption{Different number of visual tokens on video-based text generation.}
    \label{fig:short-a}
  \end{subfigure}
  \begin{subfigure}{0.45\linewidth}
    \includegraphics[scale=0.26]{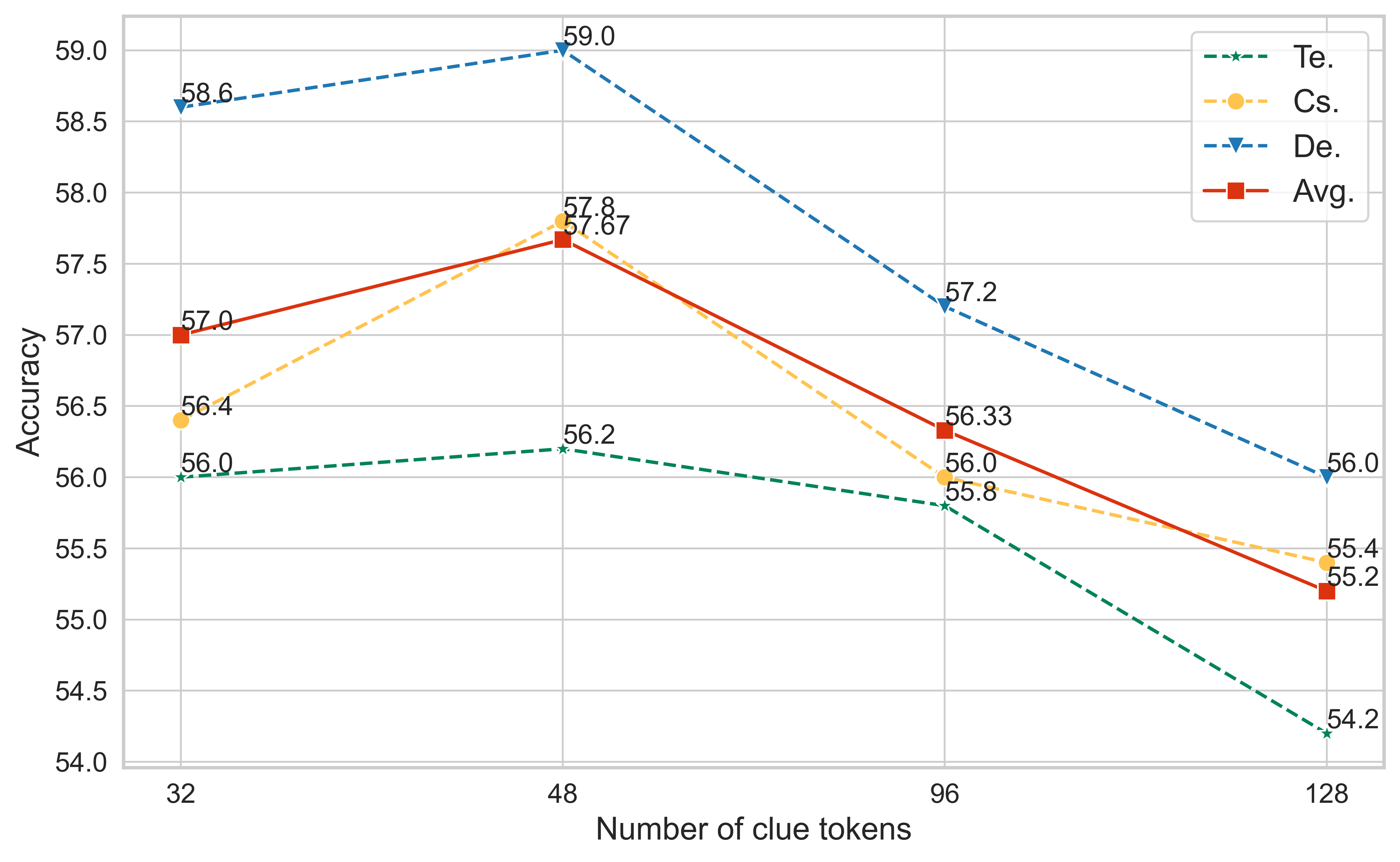}
    \caption{Different number of clue tokens on video-based text generation.}
    \label{fig:short-b}
  \end{subfigure}
  \caption{The impacts of input modal tokens. Avg. represents the average accuracy of temporal (Te.), consistency (CS.), and detail (De.).}
  \label{fig5}
   \vspace{-1.2em}
\end{figure}

\vspace{0.2em}
\noindent\textbf{Number of input modal tokens.} We evaluate the effect of the number of input modal tokens on video understanding \cite{videochatgpt}. As shown in Fig. \ref{fig5}, we first study the effect of the number of visual tokens on a time-related description task. Indeed, as the number of visual tokens increases, the accuracy in the evaluation of the various descriptions decreases, even if the change is not large, but it also confirms that the language model may not be able to reason based on longer visual tokens. Next, we examine the impact of the number of clue tokens set by the Q-former \cite{blip2}. Clue aggregator uses Q-Former to transform frame-level features into specific tokens, and we find that it works best when the length of queries $K$ is set to 48.

\vspace{0.2em}
\noindent\textbf{Impact of ADPO.} As the paper explains, supervised fine-tuning alone still underperforms in audio-visual scenarios. We test the generalizability of ADPO on Video-LLaMA \cite{videollama} and ADPO's effect on AVSD \cite{avsd} in Table \ref{tab8} to demonstrate the superiority of this learning strategy. C denotes the CiDER \cite{cider}. We find that for different MLLMs, ADPO does improve the descriptive power of the model without much learning cost, it brings different levels of enhancement to both Video-LLaMA and CAT.

\begin{figure}[t]
  \centering
  \includegraphics[scale=0.175]{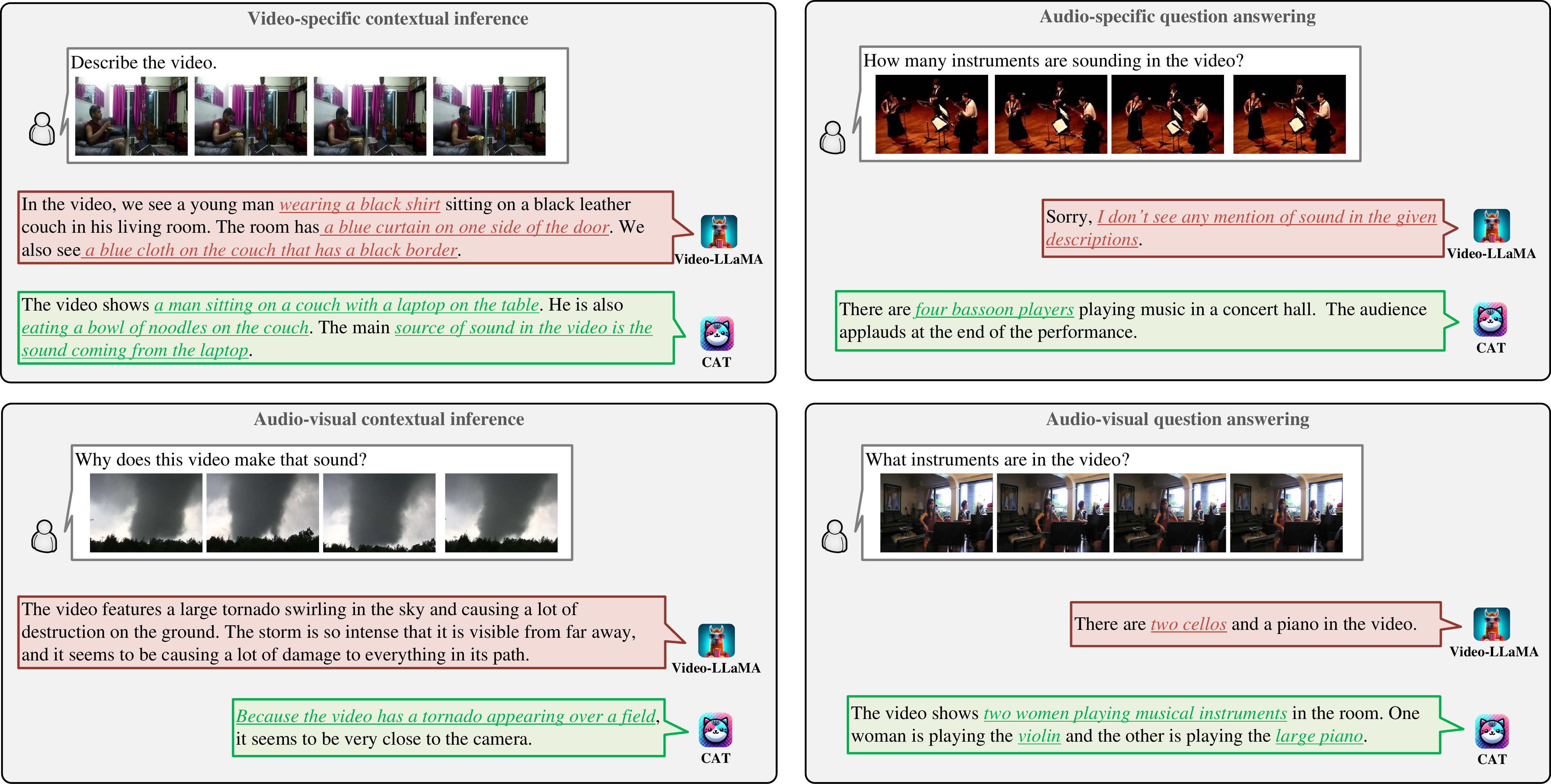}
  \vspace{-1.6em}
  \caption{Qualitative results of the video-specific contextual inference, audio-visual contextual inference, audio-specific question answering, and audio-visual question answering, respectively.
  }
   \label{instruction}
    \vspace{-1.2em}
\end{figure}

\subsection{Qualitative Analysis} 
\vspace{-0.4em}
In Fig. \ref{instruction}, we analyze the qualitative results with Video-LLaMA \cite{videollama}, which is also based on the audio-visual-language model. In the example of video-specific contextual inference, we show that CAT has a sharp perception of complex indoor scenes. Our descriptions can accurately represent what the person doing and what background sounds are in the video. Video-LLaMA is biased towards describing scene information and incorrect character information. In the specific audio question answering example, Video-LLaMA lost the ability to answer ``How many instruments are sounding in the video?''. In contrast, CAT can accurately answer the quantity question. In the audio-visual question answering example, Video-LLaMA answers incorrectly due to failed audio-visual grounding, while our CAT answers correctly due to its strong ability to capture specific objects in audio-visual scenarios.

\section{Conclusion}
In this work, we introduce CAT to enhance LLMs' multimodal understanding in dynamic audio-visual scenarios. We propose a clue aggregator to aggregate the question-related clues, which enriches the knowledge required by LLM for detailed reasoning. We mix datasets containing audio and video to empower LLM with multimodal understanding. To more consistently infer audio-visual scenarios, we collect an audio-visual joint instruction dataset to further fine-tune the CAT. Moreover, we propose AI-assisted ambiguity-aware direct preference optimization, a strategy specialized in retraining the model to eliminate ambiguous descriptions for more accurate responses to specific audio-visual objects. CAT has achieved comparable results when applied to a variety of complex audio-visual scenarios. We consider further modal extensions in the future for more comprehensive realistic applications.

\bibliographystyle{splncs04}
\bibliography{egbib}
\end{document}